\documentclass[11pt,a4paper]{article}

\usepackage[hyperref]{acl2020}
\usepackage{times}
\usepackage{latexsym}
\usepackage{graphicx}
\usepackage{xcolor}
\usepackage{amsmath}
\usepackage{array,multirow}
\usepackage{xspace}
\usepackage{lipsum}
\usepackage{enumitem}
\usepackage{qtree}
\usepackage{adjustbox}
\usepackage{dsfont}

\usepackage{titlesec}
\titlespacing*{\paragraph} {0pt}{1.25ex plus 1ex minus .2ex}{1em}

\usepackage{microtype}

\aclfinalcopy

\usepackage{algorithmicx}
\usepackage[noend]{algpseudocode}
\usepackage{algorithm}
\usepackage[T1]{fontenc}
\newcommand{\argmax}{\operatornamewithlimits{argmax}}

\algnewcommand\algorithmicdefinitions{\textbf{Definitions:}}
\algnewcommand\Definitions{\item[\algorithmicdefinitions]}

\renewcommand{\algorithmiccomment}[1]{{\color{gray}\raisebox{1px}{\texttt{\guillemotright}} #1}}

\algnewcommand{\LineComment}[1]{\Statex \hskip\ALG@thistlm \algorithmiccomment{#1}}
\algrenewcommand\alglinenumber[1]{\footnotesize #1:}
\algrenewcommand\algorithmicindent{1.0em}
\makeatletter

\newcommand{\StatexIndent}[1][3]{%
  \setlength\@tempdima{\algorithmicindent}%
  \Statex\hskip\dimexpr#1\@tempdima\relax}

\usepackage{booktabs}
\newcommand{\ra}[1]{\renewcommand{\arraystretch}{#1}}

\newcommand{\nlstring}[1]{``#1''}

\newcommand{\sentence}{\bar{x}}
\newcommand{\token}{x}
\newcommand{\embed}{\phi}

\newcommand{\tokenrep}{\textbf{\token}}

\newcommand{\vgnsl}{\textsc{VG-NSL}\xspace}
\newcommand{\combine}{{\rm combine}}
\newcommand{\score}{{\rm score}}
\newcommand{\mlp}{\textsc{MLP}}

\newcommand{\origin}{Shi2019}
\newcommand{\reproduction}{\origin$^*$}
\newcommand{\method}[3]{$#1,#2,#3$}
\newcommand{\dimone}{1}
\newcommand{\dimtwo}{2}
\newcommand{\scorews}{{\rm s}_{\mathrm{WS}}} %
\newcommand{\scoremean}{{\rm s}_\mathrm{M}} %
\newcommand{\scoremeanhi}{{\rm s}_{\mathrm{MHI}}} %
\newcommand{\combinemean}{{\rm c}_{\mathrm{ME}}} %
\newcommand{\combinemax}{{\rm c}_{\mathrm{MX}}} %

\usepackage{xr}
\makeatletter
\newcommand*{\addFileDependency}[1]{
  \typeout{(#1)}
  \@addtofilelist{#1}
  \IfFileExists{#1}{}{\typeout{No file #1.}}
}
\makeatother

\newcommand*{\myexternaldocument}[1]{
    \externaldocument{#1}
    \addFileDependency{#1.tex}
    \addFileDependency{#1.aux}
}

\newbox\jsavebox
\newcommand{\jsubfig}[2]{%
	\sbox\jsavebox{#1}%
	\parbox[t]{\wd\jsavebox}{\centering\usebox\jsavebox\\#2}%
	}

\myexternaldocument{suppl}
\myexternaldocument{main}

\title{A Critical Analysis of What is Learned \\ in Visually Grounded Neural Syntax Acquisition}

\title{What is Learned in Visually Grounded Neural Syntax Acquisition}

\author{Noriyuki Kojima, Hadar Averbuch-Elor,\\ \textbf{Alexander Rush} and \textbf{Yoav Artzi} \\
Department of Computer Science and Cornell Tech, Cornell University\\ {\tt \{nk654,he93,arush\}@cornell.edu}\\ {\tt \{yoav\}@cs.cornell.edu}\\ \\ }

\begin{document}
\maketitle

\begin{abstract}
Visual features are a promising signal for learning bootstrap textual models. However, blackbox learning models make it difficult to isolate the specific contribution of visual components. In this analysis, we consider the case study of the Visually Grounded Neural Syntax Learner~\cite{Shi2019:vgsnl}, a recent approach for learning syntax from a visual training signal. By constructing simplified versions of the model, we isolate the core factors that yield the model's strong performance. Contrary to what the model might be capable of learning, we find significantly less expressive versions produce similar predictions and perform just as well, or even better. We also find that a simple lexical signal of noun concreteness plays the main role in the model's predictions as opposed to more complex syntactic reasoning.

\end{abstract}

\section{Introduction}
\label{sec:intro}

Language analysis within visual contexts has been studied extensively, including for instruction following~\cite[e.g.,][]{Anderson:18r2r,Misra:17instructions,Misra:18goalprediction,Blukis:18drone,Blukis2019:drone-sureal}, visual question answering~\cite[e.g.,][]{Fukui:16bilinearpoolvqa, Hu:17compnetqa, Anderson:18attentionvqa}, and referring expression resolution~\cite[e.g.,][]{Mao2015:refexp,Yu:16refmscoco,Wang:16phraselocalize}.
While significant progress has been made on such tasks, the combination of vision and language makes it particularly difficult to identify what information is extracted from the visual context and how it contributes to the language understanding problem.

Recently, \citet{Shi2019:vgsnl} proposed using alignments between phrases and images as a learning signal for syntax acquisition. This task has been long-studied from a text-only setting, including recently using deep learning based approaches
\cite[inter alia]{Shen:18PRPN, Shen:18onlstm, Kim:19latent-tree-learning, Havrylov:19latent-tree-learning, Drozdov:19latent-tree-learning}. While the introduction of images provides a rich new signal for the task, it also introduces numerous challenges, such as identifying objects and analyzing scenes.

In this paper, we analyze the Visually Grounded Neural Syntax Learner (\vgnsl) model of \citet{Shi2019:vgsnl}.
In contrast to the tasks commonly studied in the intersection of vision and language, the existence of an underlying syntactic formalism allows for careful study of the contribution of the visual signal. We identify the key components of the model and design several alternatives to reduce the expressivity of the model, at times, even replacing them with simple non-parameterized rules.
This allows us to create several model variants, compare them with the full \vgnsl model, and visualize the information captured by the model parameters.

Broadly, while we would expect a parsing model to distinguish between tokens and phrases along multiple dimensions to represent different syntactic roles, we observe that the model likely does not capture such information. Our experiments show that significantly less expressive models, which are unable to capture such distinctions, learn a similar model of parsing and perform equally and even better than the original \vgnsl model. Our visualizations illustrate that the model is largely focused on acquiring a notion of noun concreteness optimized for the training data, rather than identifying higher-level syntactic roles. Our code and experiment logs are available at \url{https://github.com/lil-lab/vgnsl_analysis}.

\section{Background: VG-NSL}
\label{sec:bg}

\vgnsl consists of a greedy bottom-up parser made of three components: a token embedding function ($\embed$), a phrase combination function ($\combine$), and a decision scoring function ($\score$).
The model is trained using a reward signal computed by matching constituents and images.

Given a sentence $\sentence$ with $n$ tokens $\langle \token_1, \dots, \token_n \rangle$, the \vgnsl parser (Algorithm~\ref{alg:parser}) greedily constructs a parse tree by building up a set of constituent spans $\mathcal{T}$, which are combined spans from a candidate set $\mathcal{C}$. Parsing starts by initializing the candidate set $\mathcal{C}$ with all single-token spans. At each step, a score is computed for each pair of adjacent candidate spans $[i,k]$ and $[k+1,j]$. The best span $[i,j]$ is added to $\mathcal{T}$ and $\mathcal{C}$, and the two sub-spans are removed from $\mathcal{C}$.
The parser continues until the complete span $[1, n]$ is added to $\mathcal{T}$.

Scoring a span $[i,j]$ uses its span embedding $\tokenrep_{[i,j]}$. First, a $d$-dimensional
embedding for each single-token span is computed using $\embed$.
At each step, the score of all potential new spans $[i, j]$ are computed from the candidate embeddings $\tokenrep_{[i,k]}$ and $\tokenrep_{[k+1,j]}$.
The \vgnsl scoring function is:
\begin{equation*}
    \score(\tokenrep_{[i,k]}, \tokenrep_{[k+1,j]}) = \mlp_s([\tokenrep_{[i,k]}; \tokenrep_{[k+1,j]}])\;\;,
\end{equation*}
where $\mlp_s$ is a two-layer feed-forward network. Once the best new span is found, its span embedding is computed using a deterministic combine function.
\vgnsl computes the $d$-dimensional embedding of the span $[i,j]$ as the L2-normalized sum of the two combined sub-spans:
\begin{equation*}
    \combine(\tokenrep_{[i,k]}, \tokenrep_{[k+1,j]}) = \frac{\tokenrep_{[i,k]} + \tokenrep_{[k+1,j]}}{\left\lVert  \tokenrep_{[i,k]} +  \tokenrep_{[k+1,j]} \right\rVert_2}\;\;.
\end{equation*}

\begin{figure}
    \centering
    \vspace{-8pt}
    \begin{algorithm}[H]
    \caption{\vgnsl greedy bottom-up parser}
    \begin{algorithmic}[1]
    \footnotesize
    \Require A sentence $\sentence = \langle \token_1, \dots, \token_n\rangle$.
    \Definitions $\embed(\cdot)$ is a token embedding function; $\combine(\cdot)$ and $\score(\cdot)$ are learned functions defined in Section~\ref{sec:bg}.
    \State $\mathcal{C}, \mathcal{T} \gets \{ [i,i] \}_{i=1}^n$
    \State $\tokenrep_{[i,i]} \gets \embed(\token_i) ~~~\forall i = 1,\dots,n$
    \While{$[1,n] \notin \mathcal{T}$}
        \State $i, k, j = \displaystyle \argmax_{[i,k], [k+1,j]\in \mathcal{C}}\score(\tokenrep_{[i, k]}, \tokenrep_{[k+1, j]})$
        \State $\tokenrep_{[i,j]} \gets \combine(\tokenrep_{[i, k]}, \tokenrep_{[k+1, j]}) $
        \State $\mathcal{T} \gets \mathcal{T} \cup \{ [i,j]  \}$
        \State $\mathcal{C} \gets (\mathcal{C} \cup \{ [i,j]  \}) \setminus \{ [i,k], [k+1,j]\} $
    \EndWhile \\
    \Return $\mathcal{T}$
    \end{algorithmic}
    \label{alg:parser}
    \end{algorithm}
    \vspace{-15pt}
    \end{figure}

Learning the token embedding function $\embed$ and scoring model $\mlp_s$ relies on a visual signal from aligned images via
a reward signal derived from matching constituents and the image. The process alternates between updating the parser parameters and an external visual matching function, which is estimated by optimizing a hinge-based triplet ranking loss similar to the image-caption retrieval loss of \citet{Kiros:14vse}.
The parser parameters are estimated using a policy gradient method based on the learned visual matching function, which encourages constituents that match with the corresponding image.
This visual signal is the only objective used to learn the parser parameters. After training, the images are no longer used and the parser is text-only.

\section{Model Variations}
\label{sec:method}

We consider varying the parameterization of \vgnsl, i.e., $\phi$, $\combine$, and $\score$, while keeping the same inference algorithm and learning procedure. Our goal is to constrain model expressivity, while studying its performance and outputs.

\paragraph{Embedding Bottleneck}

We limit the information capacity of the parsing model by drastically reducing its dimensionality from $d=512$ to $1$ or $2$. We reduce dimensionality by wrapping the token embedding function with a bottleneck layer $\embed_B(\token) = \mlp_B(\embed(\token))$, where $\mlp_B$ is a two-layer feed-forward network mapping to the reduced size. This bottleneck limits the expressiveness of phrase embeddings throughout the parsing algorithm. During training, we compute both original and reduced embeddings. The original embeddings are used to compute the visual matching reward signal, whereas the reduced embeddings are used by $\score$ to determine parsing decisions. At test time, only the reduced embeddings are used.
In the case of $d=1$, the model is reduced to using a single criteria. The low dimensional embeddings are also easy to visualize, and to characterize the type of information learned.

\paragraph{Simplified Scoring}
We experiment with simplified versions of the $\score$ function.
Together with the lower-dimensional representation, this enables controlling and analyzing the type of decisions the parser is capable of.
As we control the information the embeddings can capture, simplifying the scoring function makes sure it does not introduce additional expressivity.
The first variation uses a weighted sum with parameters $\mathbf{u}, \mathbf{v}$:
\begin{equation*}
    \score_{\mathrm{WS}}(\tokenrep_{[i,k]}, \tokenrep_{[k+1,j]}) =  \mathbf{u} \cdot \tokenrep_{[i,k]} +  \mathbf{v} \cdot \tokenrep_{[k+1,j]}\;\;.
\end{equation*}
This formulation allows the model to learn structural biases, such as the head-initial (HI) bias common in English~\cite{Baker1987:atoms-language}.
The second is a non-parameterized mean, applicable for $d=1$ only:
\begin{equation*}
    \score_{\mathrm{M}}(\tokenrep_{[i,k]}, \tokenrep_{[k+1,j]}) = \frac{\tokenrep_{[i,k]} + \tau\tokenrep_{[k+1,j]}}{1 + \tau}\;\;,
\end{equation*}
where $\tau$ is a hyper-parameter that enables
upweighting the right constituent to induce a HI inductive bias. We experiment with unbiased $\tau = 1$ ($\score_{\mathrm{M}}$) and HI-biased $\tau=20$ ($\score_{\mathrm{MHI}}$) scoring.

\paragraph{Reduced Dimension Combine}
In lower dimensions, the $\combine$ function no longer
produces useful outputs, i.e., in $d=1$ it always gives $1$ or $-1$. We therefore consider mean or max pooling:
\begin{eqnarray*}
    \combine_{\mathrm{ME}}(\tokenrep_{[i,k]}, \tokenrep_{[k+1,j]})\hspace*{-0.2cm} &=&\hspace*{-0.2cm} \frac{\tokenrep_{[i,k]} +  \tokenrep_{[k+1,j]}}{2} \\
    \combine_{\mathrm{MX}}(\tokenrep_{[i,k]}, \tokenrep_{[k+1,j]}) \hspace*{-0.2cm}&=&  \\ && \hspace*{-1cm} \max(\tokenrep_{[i,k]}, \tokenrep_{[k+1,j]})\;.
\end{eqnarray*}
The mean variant computes the representation of a new span as an equal mixture of the two sub-spans, while the max directly copies to the new span representation information only from one of the spans.
The max function is similar to how head rules lexicalize parsers~\cite{Collins1996:lexaicalized}.

\section{Experimental Setup}
\label{section:setup}

We train VG-NSL and our model variants using the setup of \citet{Shi2019:vgsnl}, including
three training extensions: (a) \textbf{+HI}: adding a head-initial inductive bias to the training objective; (b) \textbf{+FastText}: the textual representations are partially initialized with pre-trained FastText~\citep{Joulin:16fasttext}; and (c) \textbf{-IN}:~\footnote{The authors of \citet{Shi2019:vgsnl} suggested this ablation  as particularly impactful on the learning outcome.} disabling the normalization of image features.
We follow the \citet{Shi2019:vgsnl} setup. We train all \vgnsl variants on 82,783 images and 413,915 captions from the MSCOCO~\cite{Lin:14mscoco} training set. We  evaluate unsupervised constituency parsing performance  using 5,000 non-overlapping held-out test captions. We use additional 5,000 non-overlapping validation captions for model selection, as well as for our analysis and visualization in Section~\ref{sec:exp}. We generate binary gold-trees using Benepar~\citep{Kitaev:18benepar}, an off-the-shelf supervised constituency parser.

We notate model variations as \method{d}{\score}{\combine}. For example, \method{\dimone}{\scorews}{\combinemean} refers to dimensionality $d=1$, weighted sum scoring function ($\scorews$), and mean pooling combine ($\combinemean$).
We train five models for each variation, and select the best checkpoint for each model by maximizing  the parse prediction agreement on the validation captions between five models. The agreement is measured by the self-$F_1$ agreement score~\cite{Williams:18latent-tree-learning-analysis}. This procedure is directly adopted from \citet{Shi2019:vgsnl}. We use the hyper-parameters from the original implementation without further tuning.

We evaluate using gold trees by reporting $F_1$ scores on the ground-truth constituents and recall on several constituent categories. We report mean and standard deviation across the five models.

\begin{table}[t]
\centering
\ra{0.95}
\setlength{\tabcolsep}{2pt}
\footnotesize
\begin{tabular}{@{}l c c c c c@{}}
\toprule
{Model} & {NP} & {VP} & {PP}  & {ADJP} & {Avg. $F_1$} \\
\midrule
\hspace{0pt} \origin & $79.6$ & $26.2$ & $42.0$ & $22.0$ & $50.4\pm0.3$\\
\hspace{0pt} \reproduction & $80.5$ & $26.9$ & $45.0 $ & $21.3 $ & $51.4\pm1.1$ \\
\hspace{0pt}\method{\dimone}{\scorews} {\combinemean} & $77.2$ & $17.0$ & $53.4$ & $18.2$ & $49.7\pm5.9 $ \\
\hspace{0pt}\method{\dimtwo}{\scorews} {\combinemean} & $\mathbf{80.8}$ & $19.1$ & $52.3$ & $17.1$ & $51.6\pm0.6 $ \\
\midrule
\textbf{+HI}\vspace{2pt}\\ 
\hspace{0pt}\origin & $74.6$ & $32.5$ & $66.5$ & $21.7$ & $53.3\pm0.2$ \\
\hspace{0pt}\reproduction & $73.1$ & $33.9$ & $64.5$ & $22.5$ & $51.8\pm0.3$ \\
\hspace{0pt}\method{\dimone}{\scorews}{ \combinemean}  & $74.0$ & $35.2$ & $62.0$ & $24.2$ & $51.8\pm0.4$ \\
\hspace{0pt}\method{\dimtwo}{\scorews}{\combinemean} & $73.8$ & $30.2$ & $63.7$ & $21.9$ & $51.3\pm0.1$ \\
\midrule
\textbf{+HI+FastText}\vspace{2pt}\\
\hspace{0pt}\origin & $78.8$ & $24.4$ & $65.6$ & $22.0$ & $54.4\pm0.3$ \\
\hspace{0pt}\reproduction & $77.3$ & $23.9$ & $64.3$ & $21.9$ & $53.3\pm0.1 $ \\
\hspace{0pt}\method{\dimone}{\scorews}{ \combinemean}&  $76.6$ & $21.9$ & $68.7$ & $20.6$ & $53.5\pm1.4$ \\
\hspace{0pt}\method{\dimtwo}{\scorews}{ \combinemean}  & $77.5$ & $22.8$ & $66.3$ & $19.3$ & $53.6\pm0.2 $ \\
\midrule
\multicolumn{2}{l}{\textbf{+HI+FastText-IN}}\vspace{2pt}\\
\hspace{0pt}\reproduction & $78.3$ & $26.6$ & $67.5$ & $22.1$ & $54.9\pm0.1 $ \\
\hspace{0pt}\method{\dimone}{\scoremean} {\combinemax} & $79.6$& $29.0$& $38.3$ & $23.5$ & $49.7\pm0.2 $ \\
\hspace{0pt}\method{\dimone}{\scoremeanhi}{\combinemax} & $77.6$& $\mathbf{45.0}$& $72.3$ & $\mathbf{24.3}$ &$\mathbf{57.5\pm0.1} $ \\
\hspace{0pt}\method{\dimone}{\scoremean}{\combinemean} & $80.0$& $26.9$ & $62.2$ & $23.2$ & $54.3\pm0.2 $ \\
\hspace{0pt}\method{\dimone}{\scoremeanhi}{\combinemean} & $76.5$ & $20.5$ & $63.6$ & $22.7$ & $52.2\pm0.3 $ \\
\hspace{0pt}\method{\dimone}{\scorews}{\combinemean} & $77.7$ & $26.3$ & $\mathbf{72.5}$ & $22.0$ & $55.5\pm0.1$ \\
\hspace{0pt}\method{\dimtwo}{\scorews}{\combinemean} & $78.5$ & $26.3$ & $69.5$ & $21.1$ & $55.2\pm0.1 $ \\
\bottomrule
\end{tabular}
\caption{Test results. We report the results from \citet{Shi2019:vgsnl} as \origin{} and our reproduction (\reproduction). 
We report mean $F_1$ and standard deviation for each system and recall for four phrasal categories.  
Our variants are specified using a representation embedding ($d\in\{1,2\}$), a $\score$ function ($\scoremean$: mean, $\scoremeanhi$: mean+HI, $\scorews$: weighted sum), and a $\combine$ function ($\combinemax$: max, $\combinemean$: mean).}
\label{tab:main_results}
\end{table}

\section{Experiments}
\label{sec:exp}

\paragraph{Quantitative Evaluation}

Table~\ref{tab:main_results} shows our main results. As the table illustrates, The model variations achieve $F_1$ scores competitive to the scores reported by \citet{Shi2019:vgsnl}  across training setups. They  achieve comparable recall on different constituent categories, and robustness to parameter initialization, quantified by self-$F_1$, which we report in an expanded version of this table in Appendix~\ref{sec:sup:resultvis}. The model variations closest to the original model, \method{\dimone}{\scorews}{\combinemean} and \method{\dimtwo}{\scorews}{\combinemean},  yield similar performance to the original model across different evaluation categories and metrics, especially in the +HI and +HI+FastText settings. Most remarkably, our simplest variants, which use $1d$ embeddings and a non-parameterized scoring function, are still competitive  (\method{\dimone}{\scoremean}{\combinemean}) or even outperform (\method{\dimone}{\scoremeanhi}{ \combinemax}) the original \vgnsl.

\begin{table}[t]
\centering
\ra{0.95}
\setlength{\tabcolsep}{2.8pt}
\footnotesize
\begin{tabular}{@{}l l c c c@{}}
\toprule
& {Training Setting} & \method{\dimone}{\scorews}{\combinemean} & \method{\dimtwo}{\scorews} {\combinemean} & {$U$}\\
\midrule
& \emph{Basic Setting} & $72.0$ & $77.5$& $87.5$\\
& \small{+HI} &$78.2$& $80.3$& $91.8$\\
& \small{+HI+FastText} &$80.5$& $83.1$& $92.3$\\
& \small{+HI+FastText-IN} &$85.6$&$86.4$& $92.8$\\
\bottomrule
\end{tabular}
\caption{Self-$F_1$ agreement between two of our variations and the original \vgnsl model. We also report the upper bound scores ($U$) calculated by directly comparing two separately trained sets of five original \vgnsl models.}
\label{tab:F_1_agreement}
\end{table}

\begin{figure}[t]
\begin{center}
\includegraphics[width=1.0\columnwidth]{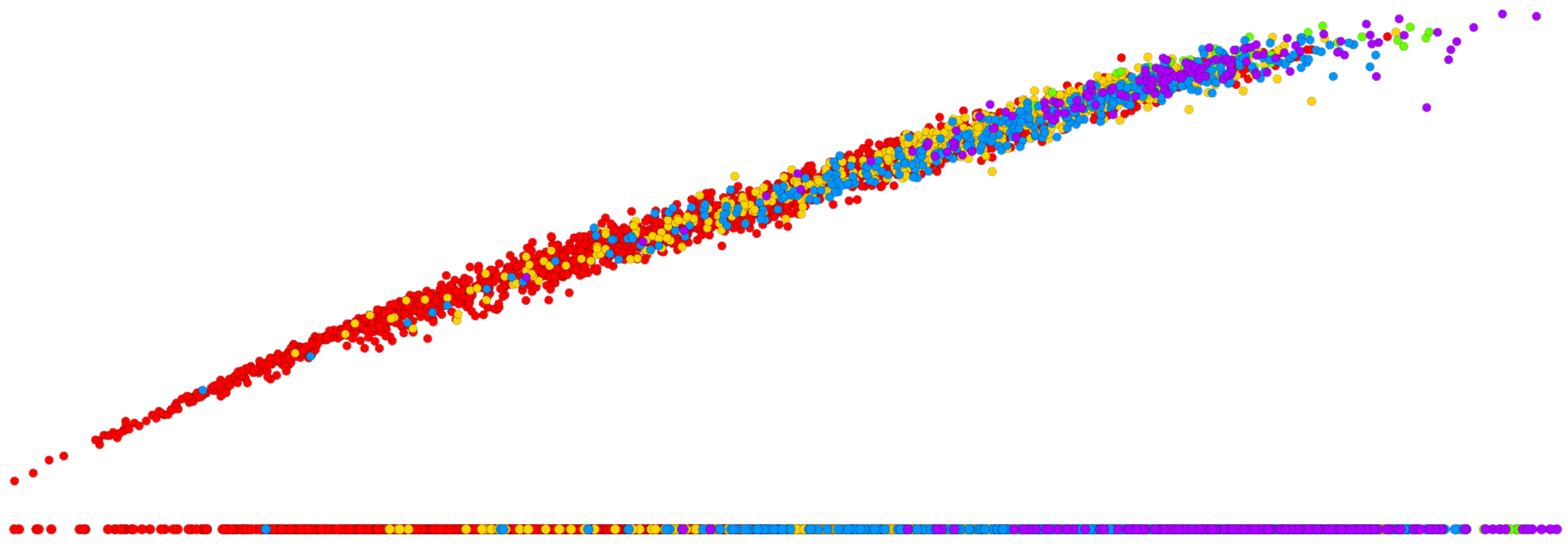} \\
\vspace{-75pt} $d=2$ \vspace{40pt} \\
$d=1$ \vspace{15pt} \\

\includegraphics[width=1.0\columnwidth]{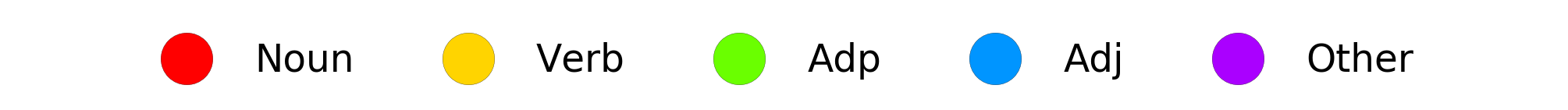}
\end{center}
\vspace{-10pt}
\caption{Token embedding visualization for  \method{\dimtwo}{\scorews}{\combinemean} (top)  and \method{\dimone}{\scorews}{\combinemean} (bottom) colored by universal POS tags~\cite{Petrov:12universalpos}.  Appendix~\ref{sec:sup:resultvis} includes an expanded version of this figure.
   }
\label{fig:token_distribution}
\end{figure}

\begin{table}[ht]
\centering
\ra{0.95}
\setlength{\tabcolsep}{2.4pt}
\footnotesize
\begin{tabular}{@{}l c c@{}|}
\hline
\toprule
Model & \method{\dimone}{\scorews}{\combinemean}\\
\midrule
\citet{Turney:11concretness} &  0.73\\
\citet{Brysbaert:14concreteness} &  0.75\\
\citet{Hessel:18concreteness} & 0.89\\
\midrule
\reproduction & 0.94 \\
\bottomrule
\end{tabular}
\caption{Pearson correlation coefficient of concreteness estimates between our \method{\dimone}{\scorews}{\combinemean} variant and existing concreteness estimates, including reproduced  estimates derived from  \vgnsl by \citet{Shi2019:vgsnl}.}
\label{tab:concretness_correlation}
\end{table}


\begin{figure}[t]

\begin{center}
    \includegraphics[width=1.0\columnwidth]{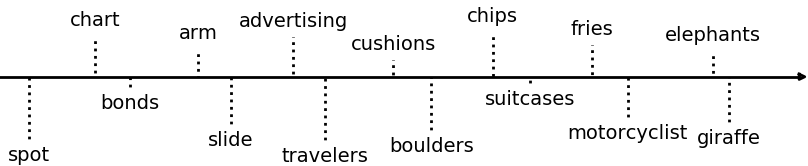}
\end{center}
\vspace{-8pt}
\caption{Noun distribution using the $1d$ representation from the \method{\dimone}{\scorews}{\combinemean} variant. The nouns are sorted by their representation value in increasing order from left.}
\label{fig:noun_distribution}
\end{figure}

\begin{table}[ht]
\centering
\ra{1}
\setlength{\tabcolsep}{2.8pt}
\footnotesize
\begin{tabular}{@{}l l c c c@{}}
\toprule
& {Training Setting} & Token  & \method{\dimone}{\scorews}{\combinemean} & \reproduction \\
\midrule
& \emph{Basic Setting} & herd & $49.5 \Rightarrow 36.3$ & $51.0 \Rightarrow 47.$6  \\
& \emph{Basic Setting*} & cat & $52.4 \Rightarrow 56.9$ & $51.0 \Rightarrow 57.2$  \\
& \small{+HI} & elephant &  $51.7 \Rightarrow 63.7$ & $51.6 \Rightarrow 59.8$\\
& \small{+HI+FastText} & motorcycle & $52.9 \Rightarrow 59.9$& $52.9 \Rightarrow 60.7$\\
& \small{+HI+FastText-IN} & elephant & $55.0 \Rightarrow 62.9$ & $54.6 \Rightarrow 60.2$\\
\bottomrule
\end{tabular}
\caption{$F_1$ scores evaluated before and after replacing nouns in captions with the most concrete token predicted by models using the \method{\dimone}{\scorews}{\combinemean} configuration. The replacement occurs during test time only as described in Section~\ref{sec:exp}. In \emph{Basic Setting}$^*$, we remove one model from \method{\dimone}{\scorews}{\combinemean} which has a significantly low $F_1$ agreement (54.2) to the rest of four models using the \method{\dimone}{\scorews}{\combinemean} configuration. }
\label{tab:replace_np}
\end{table}

Our simplified model variations largely learn the same parsing model as the original. Table~\ref{tab:F_1_agreement} shows self-$F_1$ agreement by comparing constituents predicted by our models in each training setting with the original model.
We compute this agreement measure by training two sets of five models on the training data, and selecting checkpoints using the validation captions for each of our model variants and the original \vgnsl model. We parse the same validation captions using each model and generate ten parse trees for each caption, one for each model (i.e., five for each distinct set). We calculate self-$F_1$ agreement between models by comparing parse trees from model variants to parse trees from the original \vgnsl.
We permute all 25 (five by five) combinations of variant/\vgnsl pairs and obtain self-$F_1$ agreement between the model variant and the original \vgnsl by averaging scores from each pair.
For the upper-bound agreement calculation, we train two distinct sets of five original \vgnsl models.
Our parsing model is very similar but not exactly identical: there is roughly a six points F1 agreement gap in the best case compared to the upper bound.
We consider these numbers a worst-case scenario because self-$F_1$ agreement measures on the validation data are used twice. First, for model selection to eliminate the variance of each five-model set, and second for the variant agreement analysis.

\paragraph{Expressivity Analysis}

We analyze the embeddings of the two variants closest to the original model, \method{\dimone}{\scorews}{\combinemean} and \method{\dimtwo}{\scorews}{\combinemean}, to identify the information they capture. Both behave similarly to the original \vgnsl.
Figure~\ref{fig:token_distribution} visualizes the token embedding space for these variants. Interestingly, the distribution of the $2d$ token embeddings seems almost linear, suggesting that the additional dimension is largely not utilized during learning, and that both have a strong preference for separating nouns from tokens belonging to other parts of speech. It seems only one core visual signal is used in the model and if this factor is captured, even a $1d$ model can propagate it through the tree.

We hypothesize that the core visual aspect learned, which is captured even in the $1d$ setting, is noun concreteness.
Table~\ref{tab:concretness_correlation} shows that the reduced token embeddings have strong correlations with existing estimates of concreteness.
Figure~\ref{fig:noun_distribution} shows the ordering of example nouns according to our 1$d$ learned model representation. We observe that the concreteness estimated by our model correlates with nouns that are relatively easier to ground visually in MSCOCO images. For example, nouns like \nlstring{giraffe} and \nlstring{elephant} are considered most concrete. These nouns are relatively frequent in MSCOCO (e.g., \nlstring{elephant} appears 4,633 times in the training captions) and also have a low variance in their appearances. On the other hand, nouns with high variance in images (e.g., \nlstring{traveller}) or abstract nouns (e.g., \nlstring{chart}, \nlstring{spot}) are estimated to have low concreteness.
Appendix~\ref{sec:sup:resultvis} includes examples of concreteness.

We quantify the role of concreteness-based noun identification in \vgnsl  by modifying test-time captions to replace all nouns with the most concrete token (i.e., \nlstring{elephant}), measured according to the $1d$ token embeddings learned by our model.
We pick the most concrete noun for each training configuration using mean ranking across token embeddings of the five models in each configuration.
For example, instead of parsing the original caption "girl holding a picture," we parse "elephant holding an elephant."
This uses part-of-speech information to resolve the issue where nouns with low concreteness are treated in the same manner as other part-of-speech tokens.
We compare the output tree to the original gold ones for evaluation. We observe that the $F_1$ score, averaged across the five models, significantly improves from 55.0 to 62.9 for \method{\dimone}{\scorews}{\combinemean} and from 54.6 to 60.2 for the original \vgnsl before and after our caption modification. The performance increase shows that noun identification via concreteness provides an effective parsing strategy, and further corroborates our hypothesis about what phenomena underlie the strong \citet{Shi2019:vgsnl} result. Table~\ref{tab:replace_np} includes the results for the other training settings.

\section{Conclusion and Related Work}
\label{sec:conclusion}

We studied the \vgnsl model by introducing several significantly less expressive variants, analyzing their outputs, and showing they maintain, and even improve performance.
Our analysis shows that the visual signal leads \vgnsl to rely mostly on estimates of noun concreteness, in contrast to more complex syntactic reasoning.
While our model variants are very similar to the original \vgnsl, they are not completely identical, as reflected by the self-$F_1$ scores in Table~\ref{tab:F_1_agreement}. Studying this type of difference between expressive models and their less expressive, restricted variants remains an important direction for future work. For example, this can be achieved by distilling the original model to the less expressive variants, and observing both the agreement between the models and their performance. In our case, this requires further development of distillation methods for the type of reinforcement learning setup \vgnsl uses, an effort that is beyond the scope of this paper.

Our work is related to the recent inference procedure analysis of \citet{Dyer:19onlstm-analysis}. While they study what biases a specific inference algorithm introduces to the unsupervised parsing problem, we focus on the representation induced in a grounded version of the task.
Our empirical analysis is related to \citet{Htut:18prpn-analysis}, who methodologically, and successfully replicate the results of \citet{Shen:18PRPN} to study their performance.
The issues we study generalize beyond the parsing task.
The question of what is captured by vision and language models has been studied before, including for visual question answering~\cite{Agrawal:16,Agrawal:17,Goyal:17}, referring expression resolution~\cite{Cirik2018:refexp-actually-learn}, and visual navigation~\cite{Jain2019:instruction-fidelity}. We ask this question in the setting of syntactic parsing, which allows to ground the analysis in the underlying formalism. Our conclusions are similar: multi-modal models often rely on simple signals, and do not exhibit the complex reasoning we would like them to acquire.

\section*{Acknowledgements}

Special thanks to Freda Shi for code release and prompt help in re-producing the experiments of \citet{Shi2019:vgsnl}.
This work was supported by the NSF (CRII-1656998, IIS-1901030), a Google Focused Award, and the generosity of Eric and Wendy Schmidt by recommendation of the Schmidt Futures program. We thank Jack Hessel, Forrest Davis, and the anonymous reviewers for their helpful feedback.

\bibliography{main,local}
\bibliographystyle{acl_natbib}

\clearpage
\appendix
\section{Additional Results and Visualizations}
\label{sec:sup:resultvis}
\begin{table*}[t]
\centering
\ra{0.95}
\setlength{\tabcolsep}{2pt}
\footnotesize
\begin{tabular}{@{}l c c c c c c@{}}
\toprule
{Model} & {NP} & {VP} & {PP}  & {ADJP} & {Avg. $F_1$} &  Self-$F_1$\\
\midrule
\hspace{0pt} \origin & $79.6\pm0.4$ & $26.2\pm0.4$ & $42.0\pm0.6$ & $22.0\pm0.4$ & $50.4\pm0.3$ & 87.1\\
\hspace{0pt} \reproduction & $80.5\pm1.5 $ & $26.9\pm0.9 $ & $45.0\pm2.9 $ & $21.3\pm1.2 $ & $51.4\pm1.1$ & 87.3\\
\hspace{0pt}\method{\dimone}{\scorews} {\combinemean} & $77.2\pm5.3$ & $17.0\pm5.2$ & $53.4\pm12.8$ & $18.2\pm1.0$ & $49.7\pm5.9$ & 76.0\\
\hspace{0pt}\method{\dimtwo}{\scorews} {\combinemean} & $\mathbf{80.8\pm1.1}$ & $19.1\pm1.1$ & $52.3\pm3.5$ & $17.1\pm1.0$ & $51.6\pm0.6$ & 88.1\\
\midrule
\textbf{+HI}\vspace{2pt}\\ 
\hspace{0pt}\origin & $74.6\pm0.5$ & $32.5\pm1.5$ & $66.5\pm1.2$ & $21.7\pm1.1$ & $53.3\pm0.2$ & 90.2\\
\hspace{0pt}\reproduction & $73.1\pm0.3$ & $33.9\pm0.8$ & $64.5\pm0.2$ & $22.5\pm0.4$ & $51.8\pm0.3$ & 91.6\\
\hspace{0pt}\method{\dimone}{\scorews}{ \combinemean}  & $74.0\pm0.4$ & $35.2\pm2.0$ & $62.0\pm1.1$ & $24.2\pm0.9$ & $51.8\pm0.4$ & 87.3\\
\hspace{0pt}\method{\dimtwo}{\scorews}{\combinemean}& $73.8\pm0.3$ & $30.2\pm0.4$ & $63.7\pm0.3$ & $21.9\pm0.3$ & $51.3\pm0.1$ & 93.3\\
\midrule
\textbf{+HI+FastText}\vspace{2pt}\\
\hspace{0pt}\origin & $78.8\pm0.5$ & $24.4\pm0.9$ & $65.6\pm0.1$ & $22.0\pm0.7$ & $54.4\pm0.3$ & 89.8\\
\hspace{0pt}\reproduction& $77.3\pm0.1$ & $23.9\pm0.5$ & $64.3\pm0.3$ & $21.9\pm0.3$ & $53.3\pm0.1$ & 92.2\\
\hspace{0pt}\method{\dimone}{\scorews}{ \combinemean} &  $76.6\pm0.3$ & $21.9\pm2.3$ & $68.7\pm4.1$ & $20.6\pm0.9$ & $53.5\pm1.4$ & 87.8\\
\hspace{0pt}\method{\dimtwo}{\scorews}{ \combinemean}  & $77.5\pm0.2$ & $22.8\pm0.4$ & $66.3\pm0.6$ & $19.3\pm0.7$ & $53.6\pm0.2$ & 93.6\\
\midrule
\multicolumn{2}{l}{\textbf{+HI+FastText-IN}}\vspace{2pt}\\
\hspace{0pt}\reproduction & $78.3\pm0.2$ & $26.6\pm0.3$ & $67.5\pm0.5$ & $22.1\pm1.0$ & $54.9\pm0.1$ & 92.6\\
\hspace{0pt}\method{\dimone}{\scoremean} {\combinemax} & $79.6\pm0.2$& $29.0\pm0.7$& $38.3\pm0.3$ & $23.5\pm0.6$ & $49.7\pm0.2$ & 95.5 \\
\hspace{0pt}\method{\dimone}{\scoremeanhi}{\combinemax} & $77.6\pm0.2$& $\mathbf{45.0\pm0.8}$& $72.3\pm0.2$ & $\mathbf{24.3\pm1.0}$ &$\mathbf{57.5\pm0.1}$ & 93.4\\
\hspace{0pt}\method{\dimone}{\scoremean}{\combinemean} & $80.0\pm0.2$& $26.9\pm0.2$ & $62.2\pm0.4$ & $23.2\pm0.4$ & $54.3\pm0.2$ & 95.7 \\
\hspace{0pt}\method{\dimone}{\scoremeanhi}{\combinemean} & $76.5\pm0.1$ & $20.5\pm0.8$ & $63.6\pm0.6$ & $22.7\pm0.7$ & $52.2\pm0.3$ & 94.7\\
\hspace{0pt}\method{\dimone}{\scorews}{\combinemean} & $77.7\pm0.1$ & $26.3\pm0.4$ & $\mathbf{72.5}\pm0.2$ & $22.0\pm0.6$ & $55.5\pm0.1$ & 95.5\\
\hspace{0pt}\method{\dimtwo}{\scorews}{\combinemean} & $78.5\pm0.4$ & $26.3\pm0.6$ & $69.5\pm1.2$ & $21.1\pm0.5$ & $55.2\pm0.1$ & 93.7\\
\bottomrule
\end{tabular}
\caption{Test results. We report the results from \citet{Shi2019:vgsnl} as \origin{} and our reproduction as \reproduction. We report mean $F_1$ and standard deviation for each system and mean recall and standard deviation for four phrasal categories. Our variants are specified using a representation embedding ($d\in\{1,2\}$), a $\score$ function ($\scoremean$: mean, $\scoremeanhi$: mean+HI, $\scorews$: weighted sum), and a $\combine$ function ($\combinemax$: max, $\combinemean$: mean).}
\label{tab:extended_main_results}
\end{table*}

\begin{figure*}[t]
\begin{center}
\includegraphics[width=0.96\textwidth]{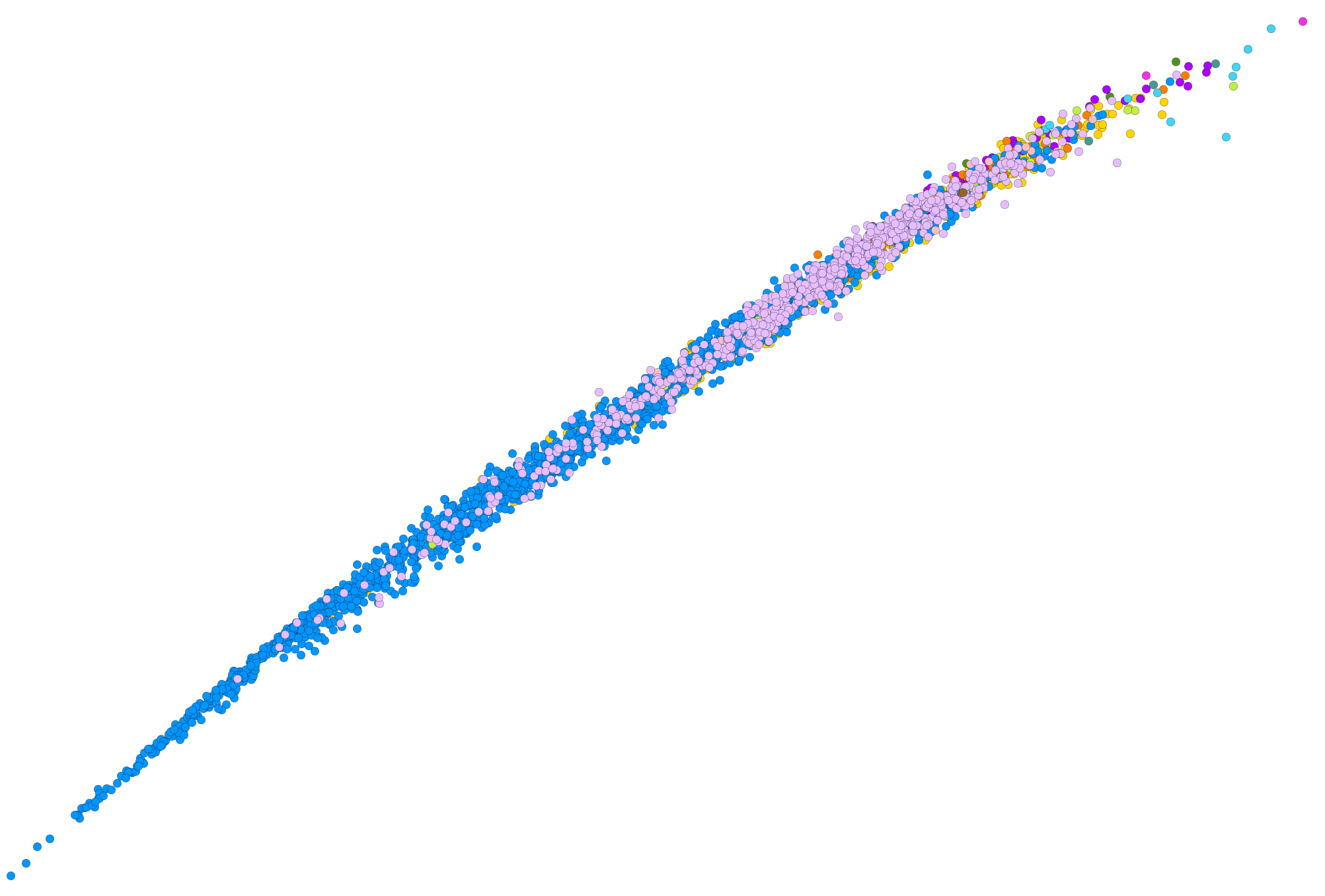} 
\includegraphics[width=0.96\textwidth]{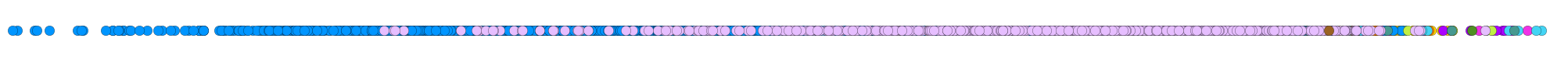} 
\includegraphics[width=0.96\textwidth]{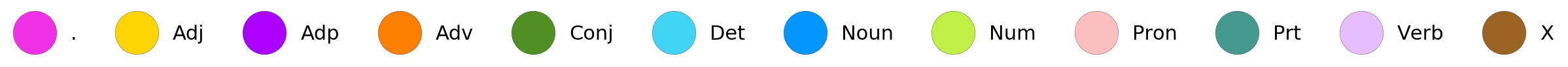} 

\end{center}
   \caption{Token embedding visualization for  \method{\dimtwo}{\scorews}{\combinemean} (top)  and \method{\dimone}{\scorews}{\combinemean} (bottom) colored by universal POS tags~\cite{Petrov:12universalpos}.}
\label{fig:token_distribution_supp}
\end{figure*}

\begin{figure*}
\footnotesize
\flushleft
		Elephant (4633 occurrences):\\
		    (a) A person riding an \textbf{elephant} and carrying gas cylinders. \\
(b) An \textbf{elephant} is in some brown grass and some trees.\\
(c) A captive \textbf{elephant} stands amid the branches of a tree in his park-like enclosure.\\
(d) Two baby gray \textbf{elephant} standing in front of each other.\\
(e) The older \textbf{elephant} is standing next to the younger \textbf{elephant}.\\
    \jsubfig{{\includegraphics[height=2.5cm]{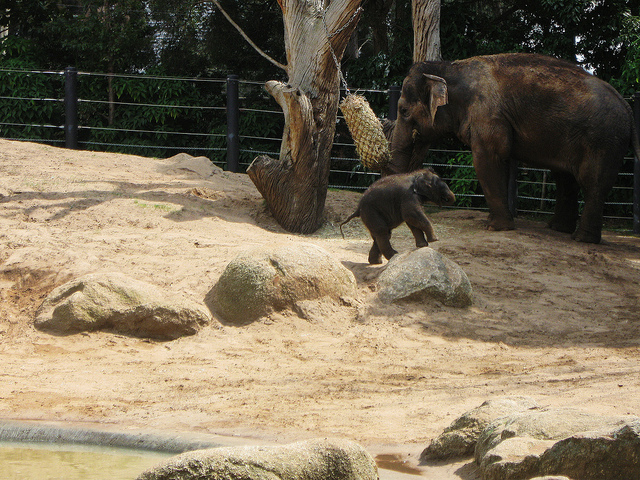}}}
	{(a)}%
    \hfill
	 \jsubfig{{\includegraphics[height=2.5cm]{{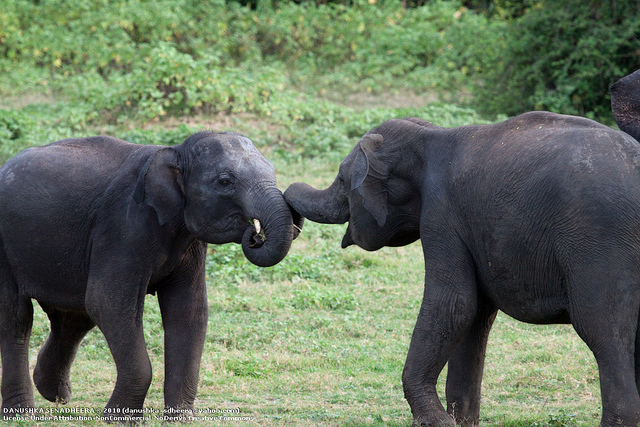}}}}
	{(b)}%
	    \hfill
	 \jsubfig{{\includegraphics[height=2.5cm]{{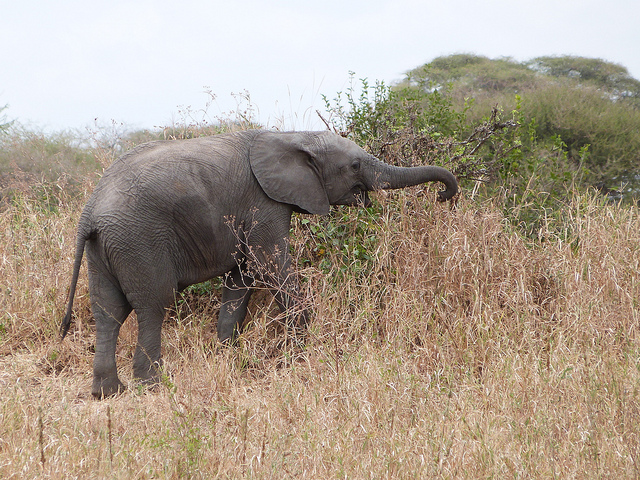}}}}
	{(c)}%
		    \hfill
	 \jsubfig{{\includegraphics[height=2.5cm]{{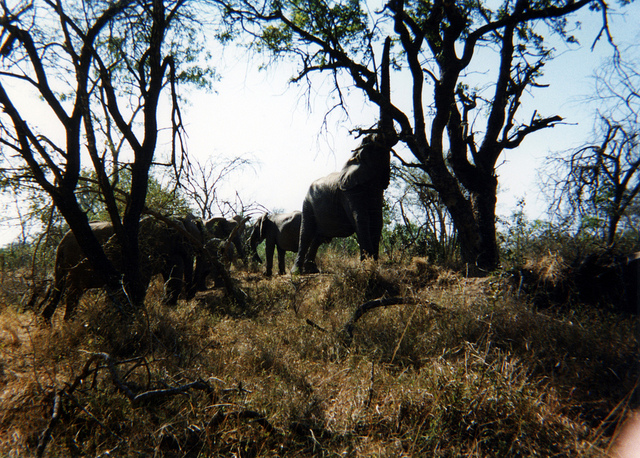}}}}
	{(d)}%
		    \hfill
	 \jsubfig{{\includegraphics[height=2.5cm]{{}}}}
	{(e)}%
	\\ \vspace{5pt}
		Giraffe (5546 occurrences): \\
		    (a) Two \textbf{giraffe} standing next to each other on a grassy field. \\
(b) A \textbf{giraffe} laying down on the dirt ground.\\
(c) A herd of \textbf{giraffe} standing next to each other on a field.\\
(d) A \textbf{giraffe} stands beneath a tree beside a marina.\\
(e) A \textbf{giraffe} rests its neck on a bunch of rocks.\\
    \jsubfig{{\includegraphics[height=2.85cm]{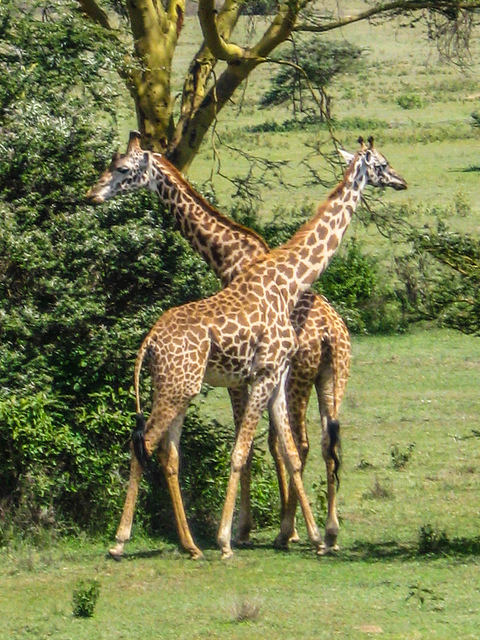}}}
	{(a)}%
    \hfill
	 \jsubfig{{\includegraphics[height=2.85cm]{{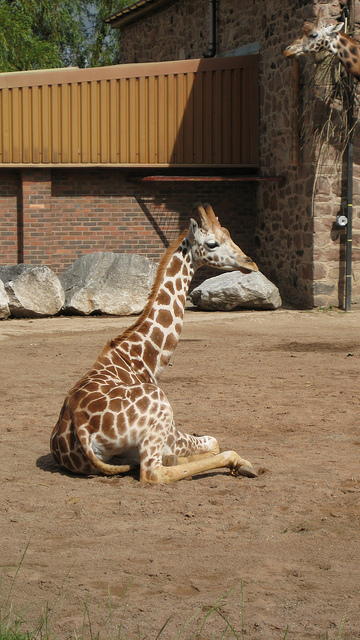}}}}
	{(b)}%
	    \hfill
	 \jsubfig{{\includegraphics[height=2.85cm]{{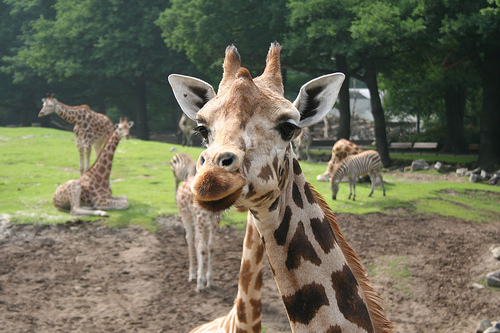}}}}
	{(c)}%
		    \hfill
	 \jsubfig{{\includegraphics[height=2.85cm]{{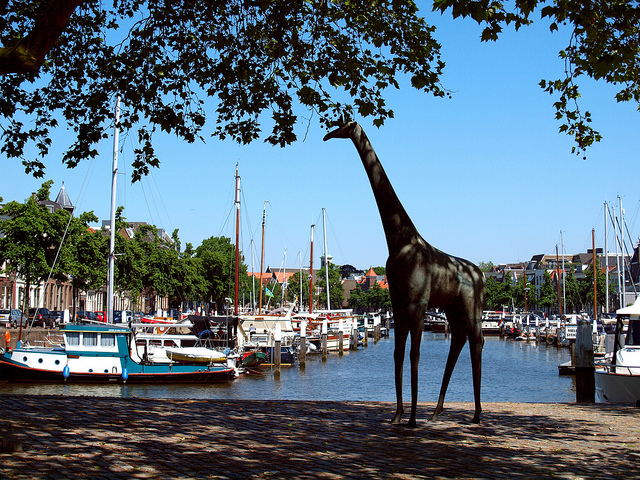}}}}
	{(d)}%
		    \hfill
	 \jsubfig{{\includegraphics[height=2.85cm]{{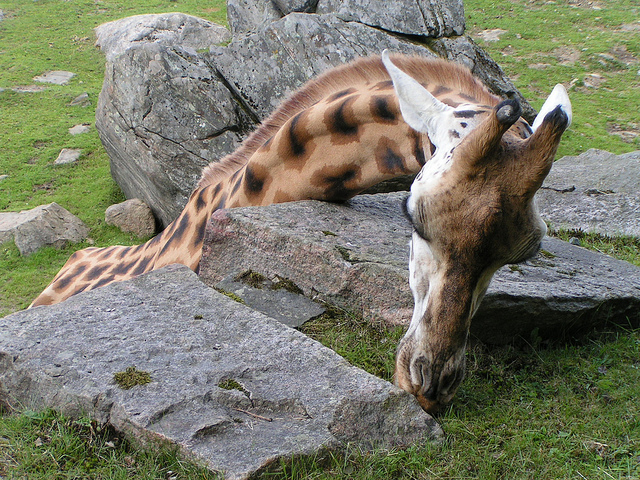}}}}
	{(e)}%
\\ \vspace{5pt}
		Pizza (8340 occurrences):\\
		    (a) A woman holding a \textbf{pizza} up in the air. \\
(b) A slice of \textbf{pizza} sitting on top of a white plate.\\
(c) A \textbf{pizza} sitting on top of a plate covered in cheese and tomatoes.\\
(d) Three pieces of sliced \textbf{pizza} on a wooden surface.\\
(e) Some boxes of frozen \textbf{pizzas} are in the store.\\
(f) A \textbf{pizza} topped with cheese and pepperoni with veggies.\\
(g) A large \textbf{pizza} is in a cardboard box.\\
    \jsubfig{{\includegraphics[height=2.2cm]{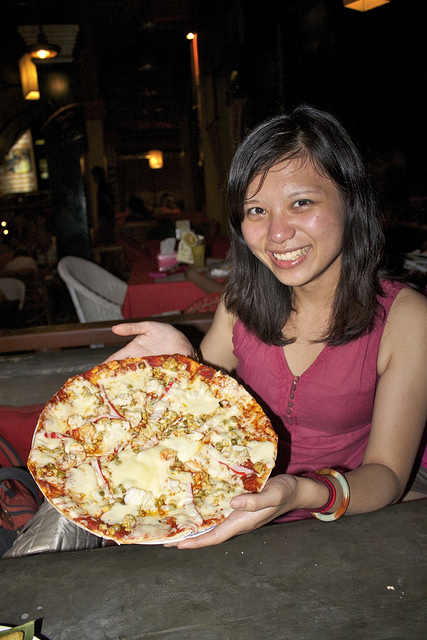}}}
	{(a)}%
    \hfill
	 \jsubfig{{\includegraphics[height=2.2cm]{{}}}}
	{(b)}%
	    \hfill
	 \jsubfig{{\includegraphics[height=2.2cm]{{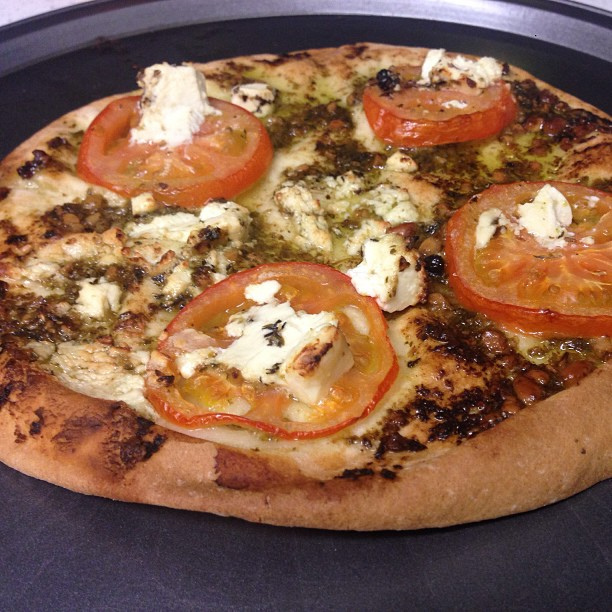}}}}
	{(c)}%
		    \hfill
	 \jsubfig{{\includegraphics[height=2.2cm]{{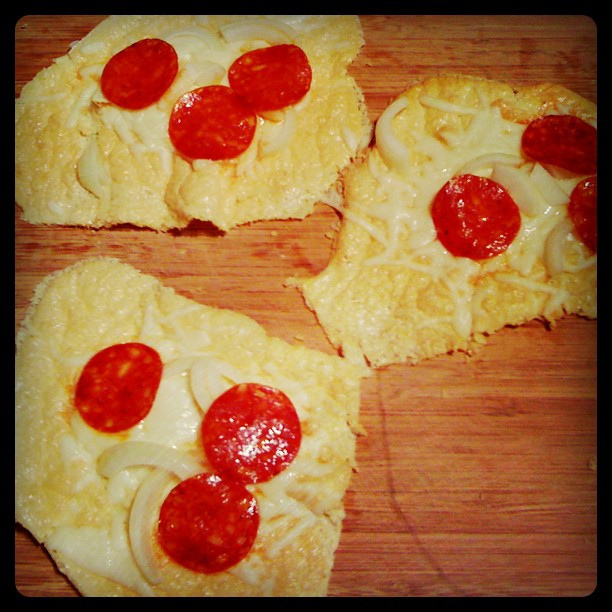}}}}
	{(d)}%
		    \hfill
	 \jsubfig{{\includegraphics[height=2.2cm]{{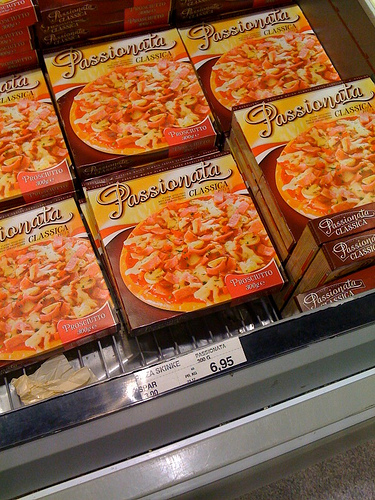}}}}
	{(e)}%
		 \hfill
	 \jsubfig{{\includegraphics[height=2.2cm]{{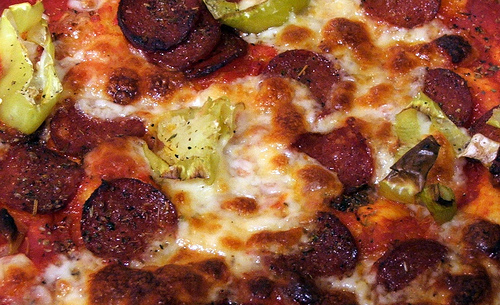}}}}
	{(f)}%
		    \hfill
	 \jsubfig{{\includegraphics[height=2.2cm]{{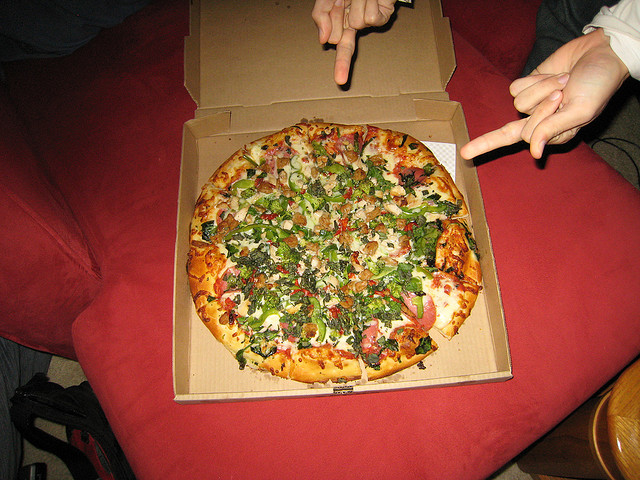}}}}
	{(g)}%
	\\ \vspace{5pt}
		Snowboarder (922 occurrences): \\
		    (a) A \textbf{snowboarder} practicing his moves at a snow facility. \\
(b) A \textbf{snowboarder} is coming down a hill and some trees.\\
(c) A \textbf{snowboarder} rests in the snow on the snowboard.\\
(d) A \textbf{snowboarder} jumps off of a hill instead of just sliding down it.\\
(e) A \textbf{snowboarder} is jumping in the air with their board held to the side.\\
(f) The snowboard is almost as big as the \textbf{snowboarder}.\\
    \jsubfig{{\includegraphics[height=2.85cm]{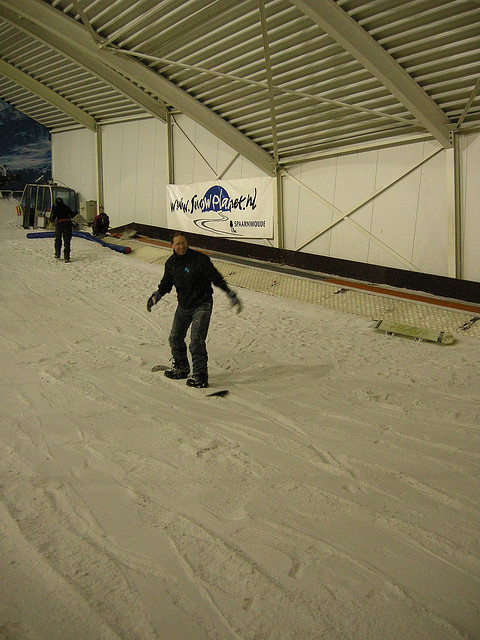}}}
	{(a)}%
    \hfill
	 \jsubfig{{\includegraphics[height=2.85cm]{{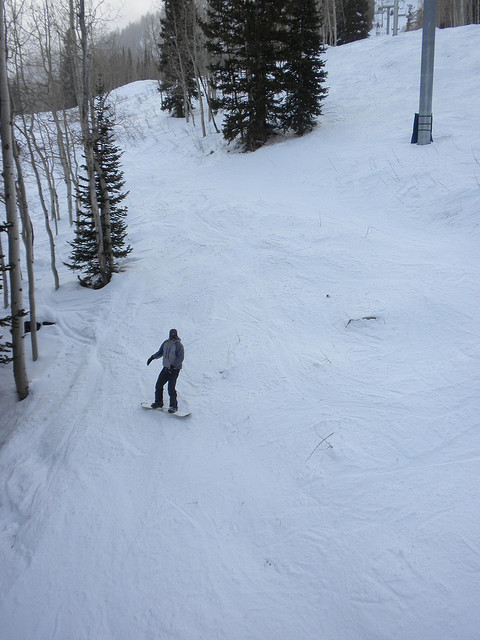}}}}
	{(b)}%
	    \hfill
	 \jsubfig{{\includegraphics[height=2.85cm]{{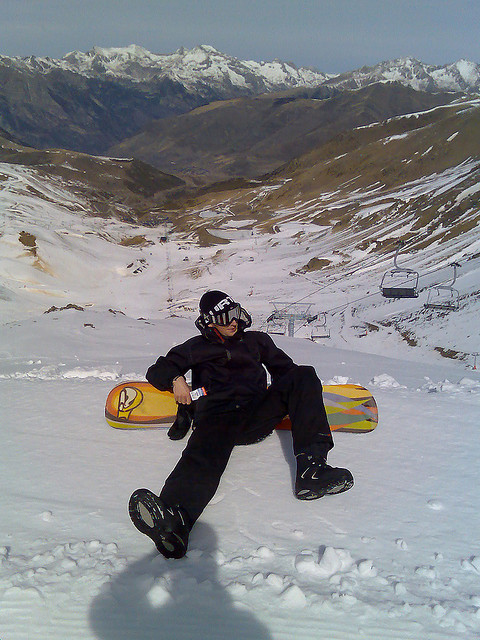}}}}
	{(c)}%
		    \hfill
	 \jsubfig{{\includegraphics[height=2.85cm]{{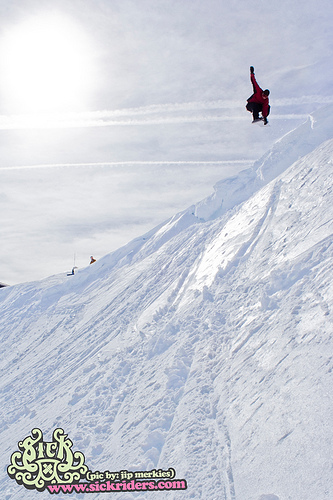}}}}
	{(d)}%
		    \hfill
	 \jsubfig{{\includegraphics[height=2.85cm]{{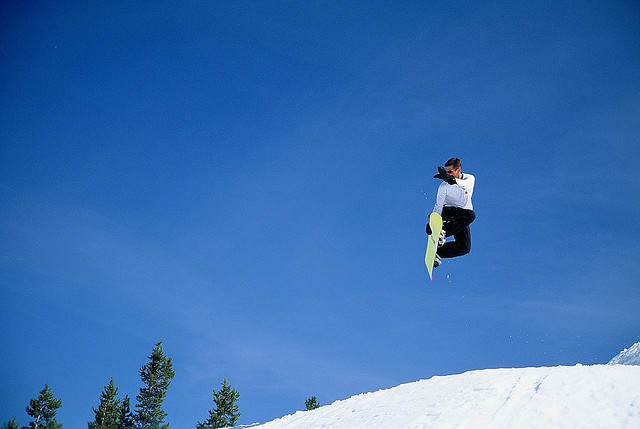}}}}
	{(e)}%
			    \hfill
	 \jsubfig{{\includegraphics[height=2.85cm]{{}}}}
	{(f)}%
\caption{Image-caption pairs corresponding to noun tokens estimated as most concrete (bottom $5\%$) in our \method{\dimone}{\scorews}{\combinemean} variant. We also report the number of occurrences in the MSCOCO training set.}
\label{fig:concreteness_high}
\end{figure*}

\begin{figure*}
\footnotesize
		\flushleft
		Metal (1630 occurrences):\\
		    (a) A pink piece of \textbf{metal} with a bolt and nut on top. \\
(b) Wilting roses and greenery in a \textbf{metal} vase.\\
(c) A couple of street signs sitting on top of a \textbf{metal} pole.\\
(d) Kitchen with wooden cabinets and a \textbf{metal} sink.\\
(e) A \textbf{metal} toilet and some tissue in a bathroom.\\
    \jsubfig{{\includegraphics[height=2.80cm]{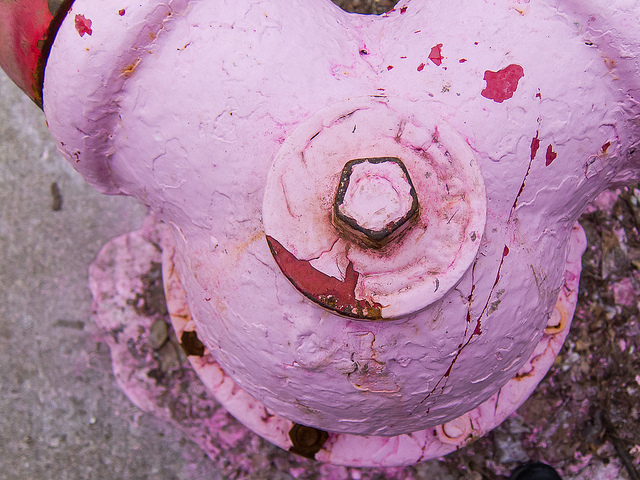}}}
	{(a)}%
    \hfill
	 \jsubfig{{\includegraphics[height=2.8cm]{{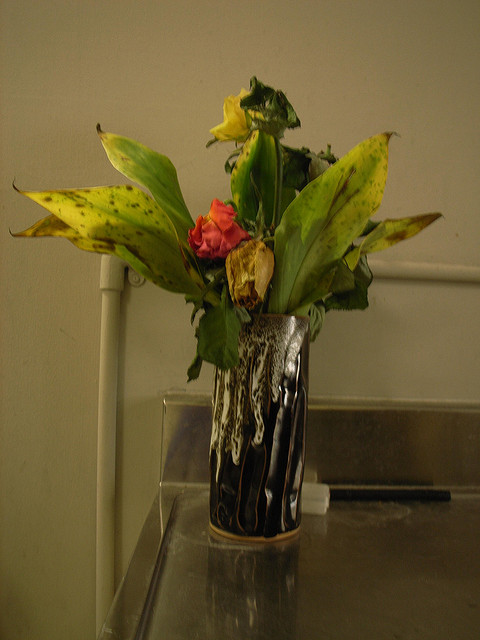}}}}
	{(b)}%
	    \hfill
	 \jsubfig{{\includegraphics[height=2.8cm]{{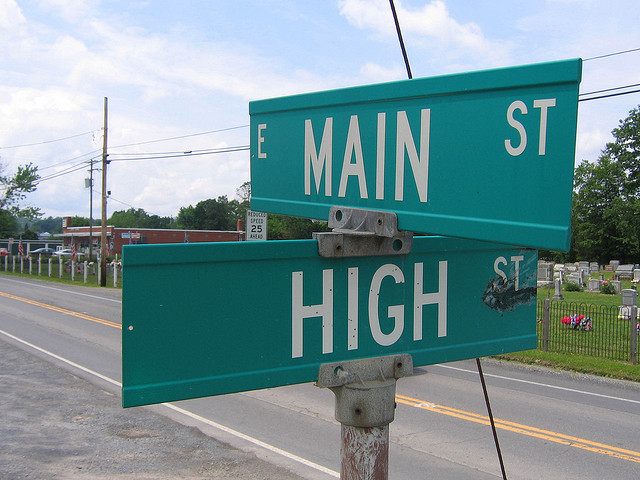}}}}
	{(c)}%
		    \hfill
	 \jsubfig{{\includegraphics[height=2.8cm]{{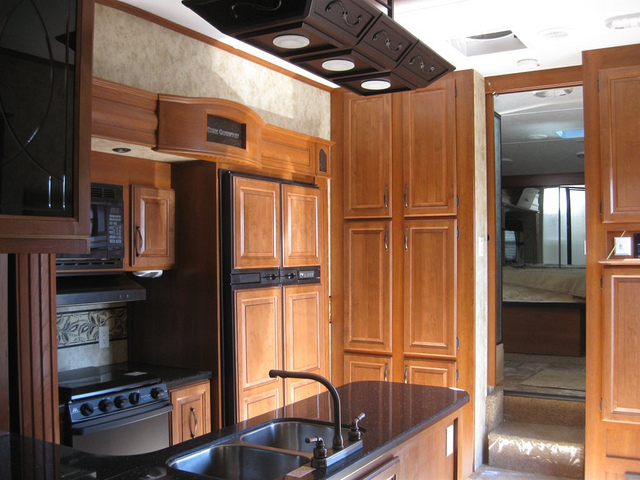}}}}
	{(d)}%
		    \hfill
	 \jsubfig{{\includegraphics[height=2.8cm]{{}}}}
	{(e)}%
	\\ \vspace{5pt}
		Palm (321 occurrences): \\
		    (a) A motorcycle sits parked in \textbf{palm} tree lined driveway. \\
(b) Two people in helmets on a parked motorcycle and a small \textbf{palm} tree to the side of them.\\
(c) Two flat bed work trucks among \textbf{palm} trees .\\
(d) A cake with \textbf{palm} trees, and a person on a surf board.\\
(e) A pink cellphone and white \textbf{palm} pilot on a table.\\
    \jsubfig{{\includegraphics[height=2.2cm]{}}}
	{(a)}%
    \hfill
	 \jsubfig{{\includegraphics[height=2.2cm]{{}}}}
	{(b)}%
	    \hfill
	 \jsubfig{{\includegraphics[height=2.2cm]{{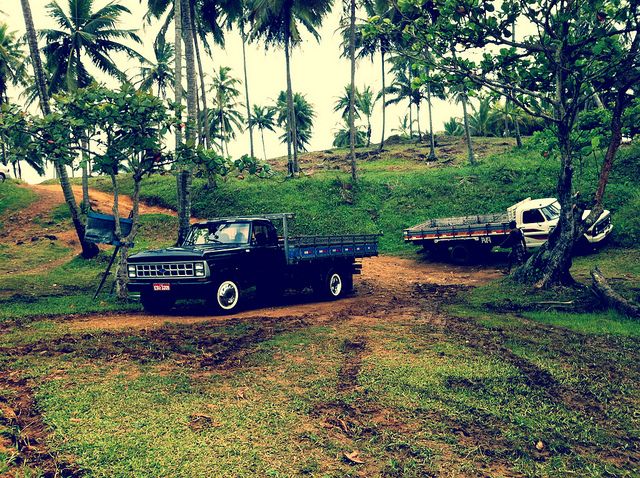}}}}
	{(c)}%
		    \hfill
	 \jsubfig{{\includegraphics[height=2.2cm]{{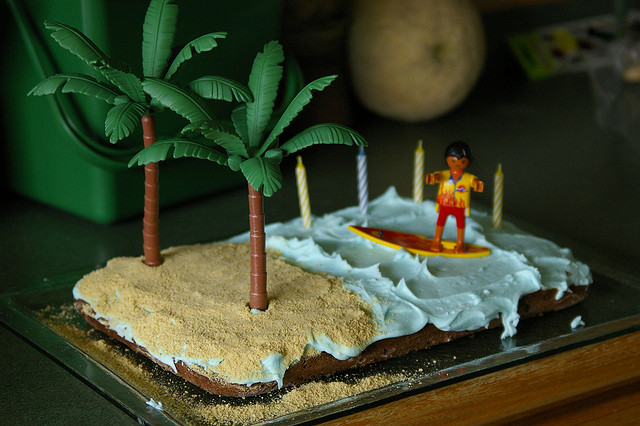}}}}
	{(d)}%
		    \hfill
	 \jsubfig{{\includegraphics[height=2.2cm]{{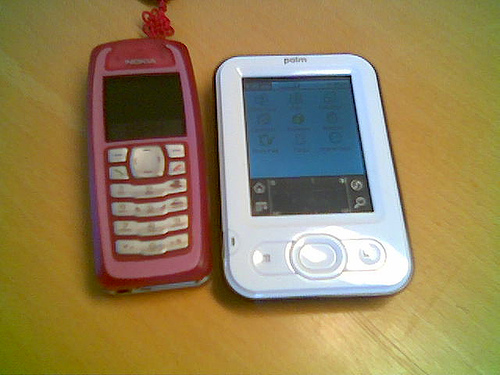}}}}
	{(e)}%
\\ \vspace{5pt}
		Picture (5932 occurrences):\\
		    (a) A blurry \textbf{picture} of a cat standing on a toilet. \\
(b) \textbf{Picture} of a church and its tall steeple.\\
(c) The street sign at the intersection of Broadway and 7th avenue is the star of this \textbf{picture}.\\
(d) A \textbf{picture} of some people playing with a frisbee.\\
(e) A little girl sitting in the middle of a restaurant and smiling for \textbf{picture}.\\

    \jsubfig{{\includegraphics[height=2.6cm]{}}}
	{(a)}%
    \hfill
	 \jsubfig{{\includegraphics[height=2.6cm]{{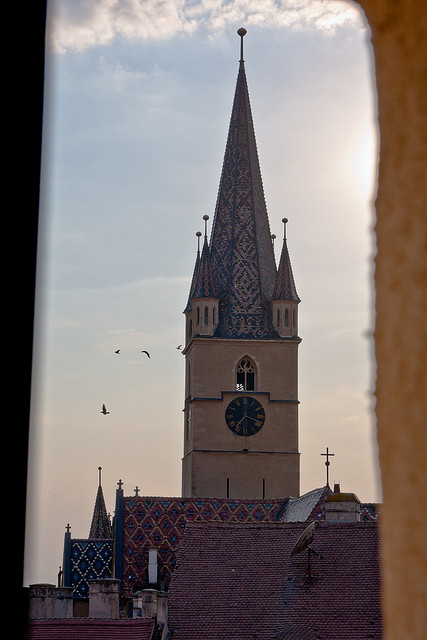}}}}
	{(b)}%
	    \hfill
	 \jsubfig{{\includegraphics[height=2.6cm]{{}}}}
	{(c)}%
		    \hfill
	 \jsubfig{{\includegraphics[height=2.6cm]{{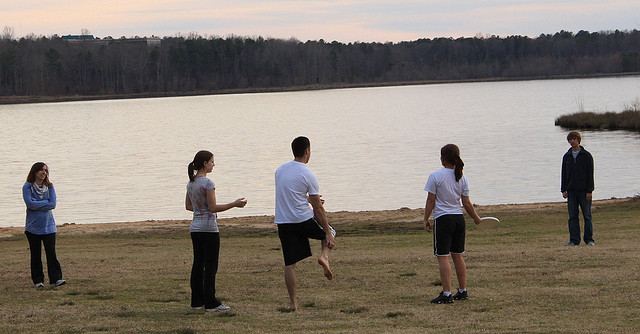}}}}
	{(d)}%
		    \hfill
	 \jsubfig{{\includegraphics[height=2.6cm]{{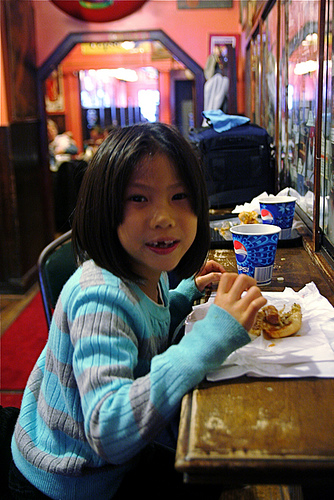}}}}
	{(e)}%
	\\ \vspace{5pt}
		Time (1184 occurrences): \\
		    (a) A \textbf{time} lapse photo of a skier skiing down a hill. \\
(b) A skaterboarder getting major air over some stairs during a night \textbf{time} shoot.\\
(c) The man is trying to eat three hot dogs are the same \textbf{time}.\\
(d) A boy playing a WII game at Christmas \textbf{time}.\\
(e) A large display of a hand holding a cell phone to tell the \textbf{time}.\\
    \jsubfig{{\includegraphics[height=2.3cm]{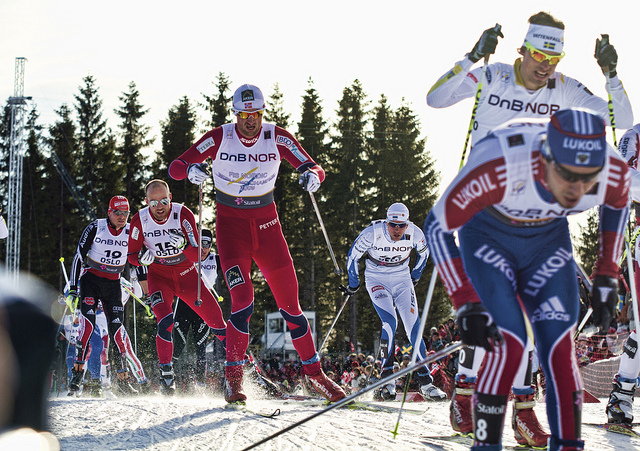}}}
	{(a)}%
    \hfill
	 \jsubfig{{\includegraphics[height=2.3cm]{{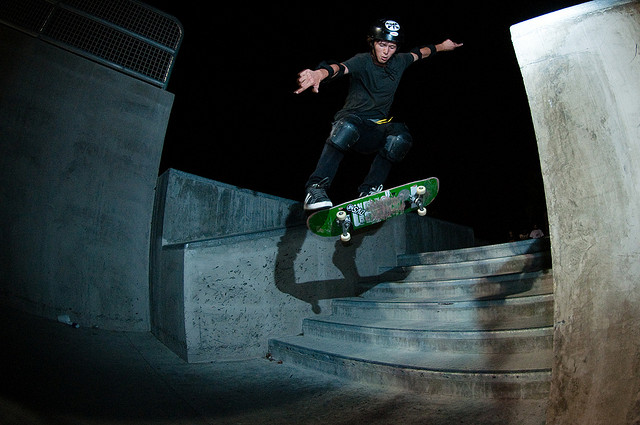}}}}
	{(b)}%
	    \hfill
	 \jsubfig{{\includegraphics[height=2.3cm]{{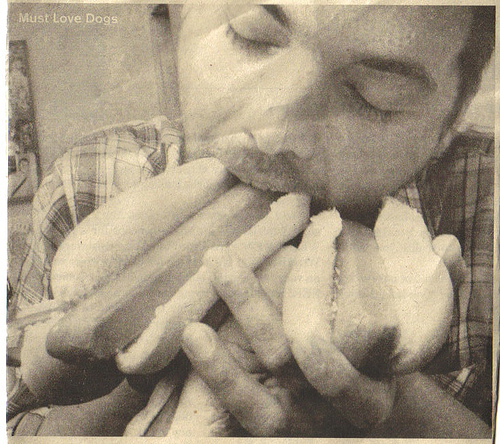}}}}
	{(c)}%
		    \hfill
	 \jsubfig{{\includegraphics[height=2.3cm]{{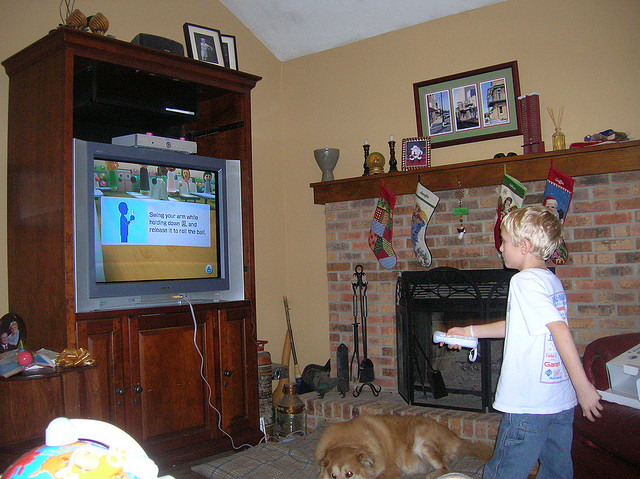}}}}
	{(d)}%
		    \hfill
	 \jsubfig{{\includegraphics[height=2.3cm]{{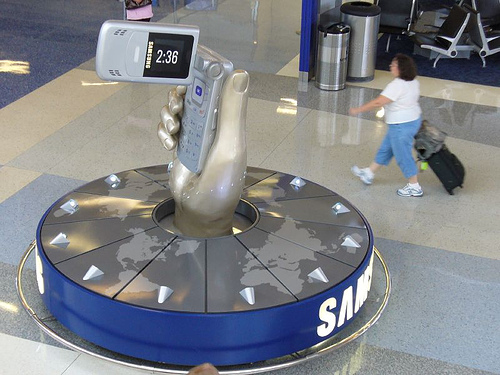}}}}
	{(e)}%
\caption{Image-caption pairs corresponding to noun tokens estimated as least concrete (bottom $5\%$) in our \method{\dimone}{\scorews}{\combinemean} variant. We also report the number of occurrences in the MSCOCO training set.}
\label{fig:concreteness_low}

\end{figure*}

Table~\ref{tab:extended_main_results} is an extended version of Table~\ref{tab:main_results} from Section~\ref{sec:exp}. We include standard deviation for the phrasal category recall and self-$F_1$ scores evaluated across different parameter initializations.
Figure~\ref{fig:token_distribution_supp} is a larger version of Figure~\ref{fig:token_distribution} from Section~\ref{sec:exp}. It visualizes the token embeddings of \method{\dimone}{\scorews}{\combinemean} and \method{\dimtwo}{\scorews}{\combinemean} for all universal parts-of-speech categories~\citep{Petrov:12universalpos}.
Figures~\ref{fig:concreteness_high} and~\ref{fig:concreteness_low} show several examples visualizing our learned representations with the \method{\dimone}{\scorews}{\combinemean} variant, the $1d$ variant closest to the original model, as a concreteness estimate. Figure~\ref{fig:concreteness_high} shows the most concrete nouns, and Figure~\ref{fig:concreteness_low} shows the least concrete nouns. We selected nouns from the top (bottom) 5\% of the data as most (least) concrete. We randomly selected image-caption pairs for these nouns.

At the end of the supplementary material, we include tree visualizations, comparing gold trees with phrasal categories,  trees generated by the original \vgnsl, and trees generated by our best performing, simplified \method{\dimone}{\scoremeanhi}{\combinemax} variant.
We select the trees to highlight the difference between \vgnsl and our variant.
First, we select all development trees where all five \vgnsl models agree to avoid results that are likely due to initialization differences.
We do the same for our variant.
Finally, we select all trees where the two sets, from \vgnsl and our variant, disagree. This process leaves us with 814 development examples, out of the original 5,000 examples. We display ten examples from this final set.

\clearpage

\clearpage
\textbf{Gold Tree}

\scriptsize

\fbox{\Tree [.S  [.NP [.DT A ] [.NN girl ] ] [.VP    [.VBZ smiles ]    [.SBAR      [.IN as ]      [.S        [.NP [.PRP she ] ]       [.VP [.VBZ holds ] [.NP [.DT a ] [.JJ kitty ] [.NN cat ] ] ] ] ] ] [.. . ] ]}

\vspace{10pt}

\textbf{VG-NSL (original)}

\fbox{\Tree [ [ a [ [ [ [ girl smiles ] [ as she ] ] holds ] [ a [ kitty cat ] ] ] ] . ]}

\vspace{10pt}

\textbf{Simplified VG-NSL ($1,s_{MHI},c_{MX}$) }

\fbox{\Tree [ [ [ a [ girl smiles ] ] [ as [ she [ holds [ a [ kitty cat ] ] ] ] ] ] . ]}

\clearpage

\textbf{Gold Tree}

\fbox{\Tree [.NP  [.NP [.DT A ] [.NN woman ] ] [.VP    [.VBG riding ]    [.NP [.DT a ] [.NN bike ] ]   [.PP      [.IN down ]      [.NP        [.NP [.DT a ] [.NN street ] ]       [.ADVP [.JJ next ] [.PP [.TO to ] [.NP [.DT a ] [.NN divider ] ] ] ] ] ] ]  [.. . ] ]}

\vspace{10pt}

\textbf{VG-NSL (original)}

\fbox{\Tree [ [ [ a [ [ woman [ riding [ a bike ] ] ] [ down [ a street ] ] ] ] [ next [ to [ a divider ] ] ] ] . ]}

\vspace{10pt}

\textbf{Simplified VG-NSL ($1,s_{MHI},c_{MX}$) }

\fbox{\Tree [ [ [ [ a [ woman [ riding [ a bike ] ] ] ] [ down [ a street ] ] ] [ next [ to [ a divider ] ] ] ] . ]}

\clearpage

\textbf{Gold Tree}

\fbox{\Tree [.NP  [.NP [.DT A ] [.NN bath ] [.NN tub ] ] [.VP    [.VBG sitting ]    [.ADVP [.JJ next ] [.PP [.TO to ] [.NP [.DT a ] [.NN sink ] ] ] ]   [.PP [.IN in ] [.NP [.DT a ] [.NN bathroom ] ] ] ] [.. . ] ]}

\vspace{10pt}

\textbf{VG-NSL (original)}

\fbox{\Tree [ [ [ [ a [ bath tub ] ] sitting ] [ next [ to [ [ a sink ] [ in [ a bathroom ] ] ] ] ] ] . ]}

\vspace{10pt}

\textbf{Simplified VG-NSL ($1,s_{MHI},c_{MX}$) }

\fbox{\Tree [ [ [ a [ bath tub ] ] [ sitting [ next [ to [ [ a sink ] [ in [ a bathroom ] ] ] ] ] ] ] . ]}

\clearpage
 \clearpage

\textbf{Gold Tree}

\fbox{\Tree [.NP  [.NP [.DT A ] [.NN parking ] [.NN meter ] ] [.PP    [.IN on ]    [.NP [.NP [.DT a ] [.NN street ] ][.PP [.IN with ] [.NP [.NNS cars ] ] ] ] ] ]}

\vspace{10pt}

\textbf{VG-NSL (original)}

\fbox{\Tree [ [ a [ parking meter ] ] [ on [ a [ street [ with cars ] ] ] ] ]}

\vspace{10pt}

\textbf{Simplified VG-NSL ($1,s_{MHI},c_{MX}$) }

\fbox{\Tree [ [ a [ parking meter ] ] [ on [ [ a street ] [ with cars ] ] ] ]}

\clearpage

\textbf{Gold Tree}

\fbox{\Tree [.NP  [.NP [.NN Bathroom ] ] [.PP    [.IN with ]    [.NP      [.NP [.DT a ] [.NX [.NN pedestal ] [.NN sink ] ] ]      [.CC and ]      [.NX [.NN claw ] [.NN foot ] [.NN bathtub ] ] ] ] ]}

\vspace{10pt}

\textbf{VG-NSL (original)}

\fbox{\Tree [ [ [ bathroom with ] [ a [ pedestal sink ] ] ] [ and [ claw [ foot bathtub ] ] ] ]}

\vspace{10pt}

\textbf{Simplified VG-NSL ($1,s_{MHI},c_{MX}$) }

\fbox{\Tree [ bathroom [ with [ [ a [ pedestal sink ] ] [ and [ claw [ foot bathtub ] ] ] ] ] ]}

\clearpage

\textbf{Gold Tree}

\fbox{\Tree [.S  [.NP [.DT A ] [.NN claw ] [.NN foot ] [.NN tub ] ] [.VP    [.VBZ is ]    [.PP      [.PP [.IN in ] [.NP [.DT a ] [.JJ large ] [.NN bathroom ] ] ]      [.PP [.IN near ] [.NP [.DT a ] [.NN pedestal ] [.NN sink ] ] ] ] ]  [.. . ] ]}

\vspace{10pt}

\textbf{VG-NSL (original)}

\fbox{\Tree [ [ [ [ a [ claw [ foot tub ] ] ] is ] [ in [ [ a [ large bathroom ] ] [ near [ a [ pedestal sink ] ] ] ] ] ] . ]}

\vspace{10pt}

\textbf{Simplified VG-NSL ($1,s_{MHI},c_{MX}$) }

\fbox{\Tree [ [ [ [ a [ claw [ foot tub ] ] ] [ is [ in [ a [ large bathroom ] ] ] ] ] [ near [ a [ pedestal sink ] ] ] ] . ]}

\clearpage

\textbf{Gold Tree}

\fbox{\Tree [.NP  [.NP [.NN Picture ] ] [.PP    [.IN of ]    [.NP      [.NP [.DT a ] [.NN church ] ]     [.CC and ]      [.NP [.PRP\$ its ] [.JJ tall ] [.NN steeple ] ] ] ] [.. . ] ]}

\vspace{10pt}

\textbf{VG-NSL (original)}

\fbox{\Tree [ [ [ picture [ of [ a church ] ] ] [ and [ its [ tall steeple ] ] ] ] . ]}

\vspace{10pt}

\textbf{Simplified VG-NSL ($1,s_{MHI},c_{MX}$) }

\fbox{\Tree [ [ picture [ of [ [ a church ] [ and [ its [ tall steeple ] ] ] ] ] ] . ]}

\clearpage

\textbf{Gold Tree}

\fbox{\Tree [.NP  [.NP [.DT A ] [.NNP giraffe ] ] [.VP    [.VBG laying ]    [.PRT [.RP down ] ]   [.PP [.IN on ] [.NP [.DT the ] [.NN dirt ] [.NN ground ] ] ] ] [.. . ] ]}

\vspace{10pt}

\textbf{VG-NSL (original)}

\fbox{\Tree [ [ [ [ [ a giraffe ] laying ] down ] [ on [ [ the dirt ] ground ] ] ] . ]}

\vspace{10pt}

\textbf{Simplified VG-NSL ($1,s_{MHI},c_{MX}$) }

\fbox{\Tree [ [ [ [ a giraffe ] laying ] [ down [ on [ [ the dirt ] ground ] ] ] ] . ]}

\clearpage

\textbf{Gold Tree}

\fbox{\Tree [.NP  [.NP [.DT A ] [.NN fire ] [.NN hydrant ] ] [.SBAR    [.WHNP [.WDT that ] ]   [.S      [.VP        [.VBZ is ]        [.VP [.VBN painted ] [.S [.ADJP [.JJ yellow ] [.CC and ] [.JJ blue ] ] ] ] ] ] ]  [.. . ] ]}

\textbf{VG-NSL (original)}

\fbox{\Tree [ [ [ a yellow ] [ and [ [ blue [ fire hydrant ] ] [ on [ the sidewalk ] ] ] ] ] . ]}

\vspace{10pt}

\textbf{Simplified VG-NSL ($1,s_{MHI},c_{MX}$) }

\fbox{\Tree [ [ [ [ a yellow ] [ and [ blue [ fire hydrant ] ] ] ] [ on [ the sidewalk ] ] ] . ]}

\clearpage

\textbf{Gold Tree}

\fbox{\Tree [.S  [.NP [.NP [.DT A ] [.NN bird ] ][.PP [.IN with ] [.NP [.JJ red ] [.NNS eyes ] ] ] ] [.VP    [.VBN perched ]    [.PP      [.IN on ]      [.NP        [.NP [.NN top ] ]       [.PP [.IN of ] [.NP [.DT a ] [.NN tree ] [.NN branch ] ] ] ] ] ] [.. . ] ]}

\vspace{10pt}

\textbf{VG-NSL (original)}

\fbox{\Tree [ [ [ [ [ a bird ] [ with [ red [ eyes perched ] ] ] ] [ on top ] ] [ of [ a [ tree branch ] ] ] ] . ]}

\vspace{10pt}

\textbf{Simplified VG-NSL ($1,s_{MHI},c_{MX}$) }

\fbox{\Tree [ [ [ [ a bird ] [ with [ red [ eyes perched ] ] ] ] [ on [ top [ of [ a [ tree branch ] ] ] ] ] ] . ]}

\clearpage

\end{document}